# Knowledge Graph Enhanced Aspect-Level Sentiment Analysis

Kavita Sharma, Ritu Patel, Sunita Iyer
Department of Computer Science, Banaras Hindu University
{Kavita.Sharma, Ritu.Patel, ISunita}@bhu.ac.in

**Abstract**

In this paper, we propose a novel method to enhance sentiment analysis by addressing the challenge of context-specific word meanings. It combines the advantages of a BERT model with a knowledge graph based synonym data. This synergy leverages a dynamic attention mechanism to develop a knowledge-driven state vector. For classifying sentiments linked to specific aspects, the approach constructs a memory bank integrating positional data. The data are then analyzed using a DCGRU to pinpoint sentiment characteristics related to specific aspect terms. Experiments on three widely used datasets demonstrate the superior performance of our method in sentiment classification.

**Keywords** Aspect-Level Sentiment Analysis, Knowledge Graph, Attention Mechanism, Deep Learning

With the rapid development of the Internet, there is an explosive growth in the number of users expressing their comments on topics in social media, and the flow of large amounts of information affects the decision-making process of organizations, and the study of these comment texts contains great value [1]. Perspective-level text sentiment analysis [2] aims to determine the sentiment polarity of each perspective word expressed in the text, in order to provide a more comprehensive, in-depth and fine-grained sentiment analysis than the document level or sentence level, to provide people with a convenient and automated tool to improve the utilization of information on the Internet.

The methods for perspective-level text sentiment classification problem usually extract and learn text features for constructing classification models, which mainly include sentiment lexicon based methods and machine learning based methods. Liu [3] rated words or phrases containing emotions and solved the problem of emotion transfer due to negatives or inflections. Moghaddam et al. [4] extracted important information about a product and rated it with stars to determine the level of user satisfaction. Some researchers also use SVM classifier [5], Ngram [6] and so on to determine the emotional polarity. Although the above methods have achieved some results, they require a lot of manpower, material and resources to design semantic and syntactic features, and the performance of the methods depends on these features to a large extent, and the generalization ability is poor.

Deep learning methods usually automatically obtain low-dimensional text features from perspective words and text, and construct neural network models for perspective-level text sentiment analysis, mainly including neural network models and attention mechanism models [7]. Dong et al. [8] proposed the Target Dependent Long Short Term Memory (TDLSTM), an adaptive recurrent neural network modeling the adaptive propagation of emotion words to specific perspective words, which depends entirely on the grammatical dependency tree, but may not work because of non-standard text. The process depends entirely on the syntactic dependency tree, but may not work due to non-standard text.Tang et al. [9] divided sentences into left and right parts of perspective words and used two Long Short Term Memory (LSTM) networks to model the correlation between perspective words and their left and right contexts, respectively. Huang et al. [10] used parameterized filters and threshold mechanisms to incorporate phase information into convolutional neural networks to effectively capture specific textual features. Lei et al. [11] proposed a semantic cognitive network to simulate the cognitive process of human reading and better capture the relationship between context and perspective words.In the above methods, each word is equally important to the categorization result, and the different degree of contribution of words to the sentiment categorization in different perspectives has not been considered. In addition, we might consider logical rules [12-15] for knolwegde graph reasoning. Various models like ATABA, ATAE-LSTM, MemNet, IANs, RAM [16-17] introduced attention mechanisms to address the issue of treating all words equally in neural networks. They consider contextual words and perspective words to improve text understanding but often overlook the varying semantic information of words in different contexts, resulting in reduced prediction performance.

To address the above problems, this paper proposes a perspective-level sentiment analysis method based on Knowledge Graph and DCGRU. Firstly, we learn the contextual relationship between words in text through BERT, and design the dynamic attention mechanism to integrate the process of Knowledge Graph: combining the current word vector representation with the synonym vector representation in the Knowledge Graph to construct the memory content, and introducing the Sentinel Vector to avoid the misleading external knowledge, to get the semantic information in the current context, to further improve the text classification and to enhance the semantic information of text. We also introduce sentinel vectors to avoid external knowledge misdirection, obtain semantic information that fits the current context, and further improve the prediction performance of text classification. In this paper, we use the Twitter dataset collected by Laptop, Restaurant, and Dong [8], which is a public dataset of SemEval 2014 Task4, to conduct experiments to validate the effectiveness of this paper's method on the perspective-level text sentiment analysis problem.

## 1 A Perspective-Level Sentiment Analysis Method

### 1.1 Problem Formalization

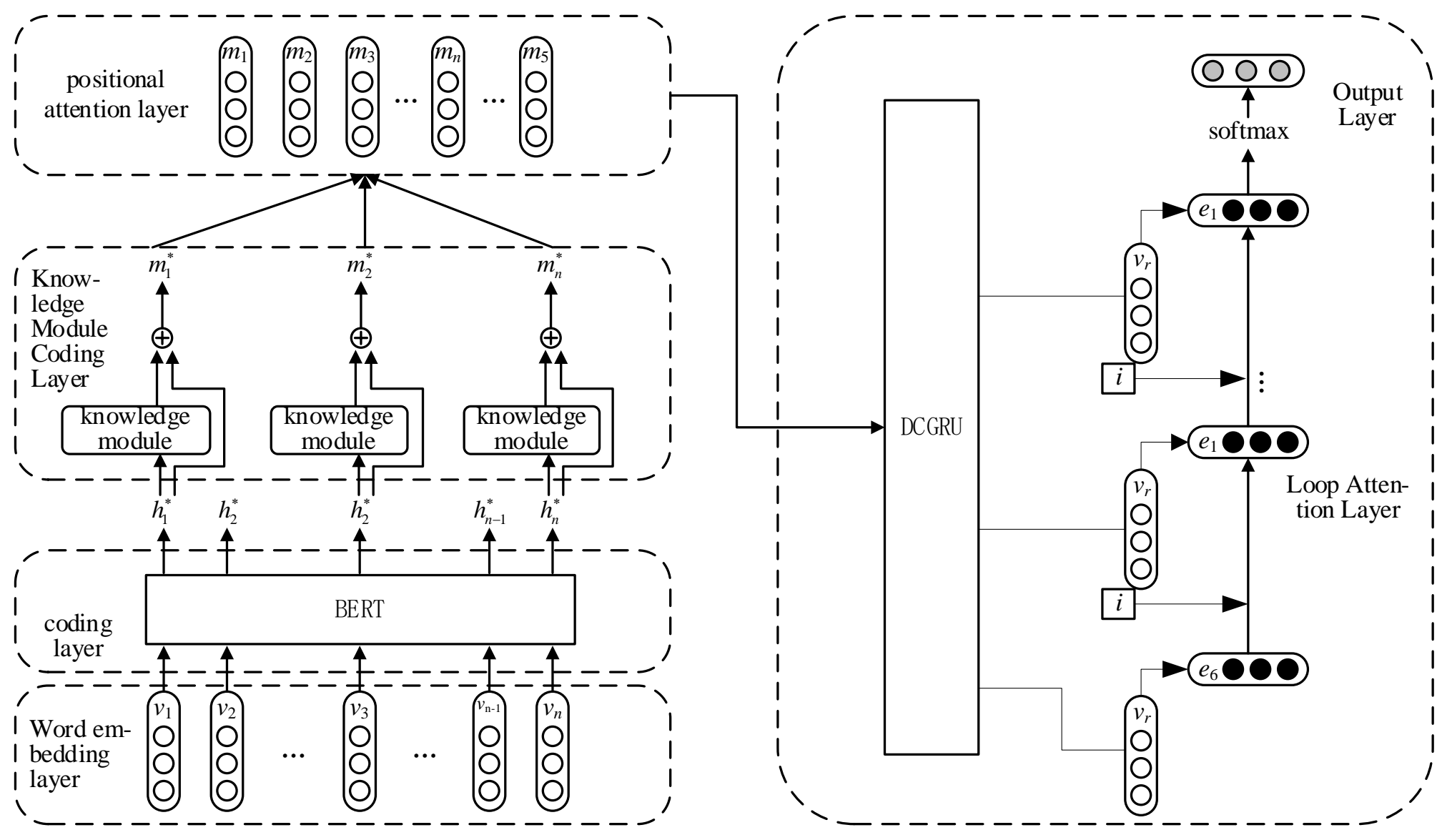

Fig.1 Framework of our aspect level sentiment analysis model

The task of perspective-level text sentiment categorization aims to learn a classifier that takes a text and a perspective word from the text and inputs it into the classifier, and determines the text's sentiment category label with respect to the given perspective.

This is essentially a three-classification problem. In this paper, the problem is formally defined as follows: Assume that given a text

$$\boldsymbol{S} = \{\boldsymbol{w}_1, \boldsymbol{w}_2, \cdots, \boldsymbol{w}_i, \boldsymbol{w}_{i+1}, \cdots, \boldsymbol{w}_{i+l}, \cdots, \boldsymbol{w}_N\}$$

where the perspective words in the text are

$$\boldsymbol{w_a} = \{\boldsymbol{w_i}, \boldsymbol{w_{i+1}}, \cdots, \boldsymbol{w_{i+l}}\}$$

Category set $y = \{y_1, y_2, y_3\}$. The goal of the sentiment classification task of perspective text is to learn a classification model to map text $S$ and perspective word $\boldsymbol{w}_a$ to corresponding category label $y_i$, that is,

$$\boldsymbol{f}(\boldsymbol{S}, \boldsymbol{w_a}) \rightarrow \boldsymbol{y_i}$$

Among them, the text $S$ contains $N$ words, and the perspective word $\boldsymbol{w}_a$ contains $l+1$ word category sets $y_1$、$y_2$、$y_3$, which respectively indicate that the emotional polarity is positive, neutral and negative.

1.2 Method components

The block diagram of the perspective-level text sentiment categorization method is shown in Figure 1. The method consists of 6 components. 1) Input text word embedding layer: the text is word-vectorized. 2) A word encoding layer: capturing semantic dependencies. 3) Coding layer fusing knowledge modules: modeling the synonyms and text information in the knowledge graph. 4) Text representation incorporating positional attention information: modeling the positional information of words and calculating the contribution of each word to the sentiment classification of the text. 5) Perspective-level affective feature representation based on DCGRU: Calculate the attention score of each memory content, and extract textual features through the threshold control unit. 6) Output layer: output the sentiment classification result of the text.

1.2.1 Input text word embedding layer

The input of this method consists of two parts. 1) Sentence text sequence

$$\boldsymbol{S} = \{\boldsymbol{w_1}, \boldsymbol{w_2}, \cdots, \boldsymbol{w_i}, \boldsymbol{w_{i+1}}, \cdots, \boldsymbol{w_{i+l}}, \cdots, \boldsymbol{w_N}\};$$

2) Sequence of perspective words appearing in the sentence

$$\boldsymbol{w_a} = \{\boldsymbol{w_i}, \boldsymbol{w_{i+1}}, \cdots, \boldsymbol{w_{i+l}}\}.$$

Among them, the perspective word $\boldsymbol{w}_a$ is a subsequence of sentence $S$, and the perspective word can be one word or multiple words. Each word $\boldsymbol{w}_i \in \mathbf{R}'^{n}$, $|V|$ is the vocabulary size. Suppose $\boldsymbol{L} \in \mathbf{R}^{d\times 1\eta}$ represents the pre-training word vector search generated by Global Vector, GloVe). In this paper, each word $\boldsymbol{w}_i$ is mapped into a corresponding word vector representation $v_i \in \mathbf{R}^{d\times 1}$ through the word embedding matrix $L$. The word embedding representation of the $S$-mapped sentence text sequence input is as follows

$$\boldsymbol{V_s} = \{\boldsymbol{v_1}, \boldsymbol{v_2}, \cdots, \boldsymbol{v_i}, \boldsymbol{v_{i+l}}, \cdots, \boldsymbol{v_N}\}$$

The word embedding of the perspective word sequence $\boldsymbol{w}_a$ is expressed as

$$\boldsymbol{V_a} = \{\boldsymbol{v_i}, \boldsymbol{v_{i+1}}, \cdots, \boldsymbol{v_{i+l}}\}$$

If the perspective word is a single word, take the word embedding vector as the representation of the perspective word sequence, if the perspective word is composed of multiple words, take the average of the word embedding vectors of multiple words as the representation of the perspective word sequence.

1.2.2 Word coding layer

The Transformer classification model constructed in this paper is based on the Encoder structure. Due to the large number of training parameters and complexity of the Transformer model, only two encoders are used, and each encoder contains a multi-head attention sublayer and a feed-forward network sublayer. The output size of all sublayers in the model as well as the embedded layers is 200. Sentence Embedding and Position Embedding are added as inputs, the first sublayer of the encoder is Multihead Attention, and the outputs are represented as sublayer (x), after residual linkage and layer norm (Add & Layer Norm, LN) the output is.

$$\text{output} = \text{LN}\ (x + \text{sublayer}\ (x))$$

The second sublayer of the encoder is the itemized Feed Forward Network (FFN), which consists of two linear transformations, where each layer has different parameters, the dimensions of the inputs and outputs are 200, and the dimensions of the internal layer are 2400 .

$$\text{FFN ( output )} = \text{Max}\ (0, \text{output} * \boldsymbol{W}_1 + b_1)\boldsymbol{W}_2 + b_2$$

Where $W_1$、$W_2$ and $b_1$、$b_2$ are the weights and deviations of the corresponding linear transformation of output, respectively. After the linear transformation, the final output of the second sublayer of the last encoder is passed into the fully connected layer after residual concatenation and layer normalization, and the classification result is output. The training process uses the Cross Entropy Loss loss function.

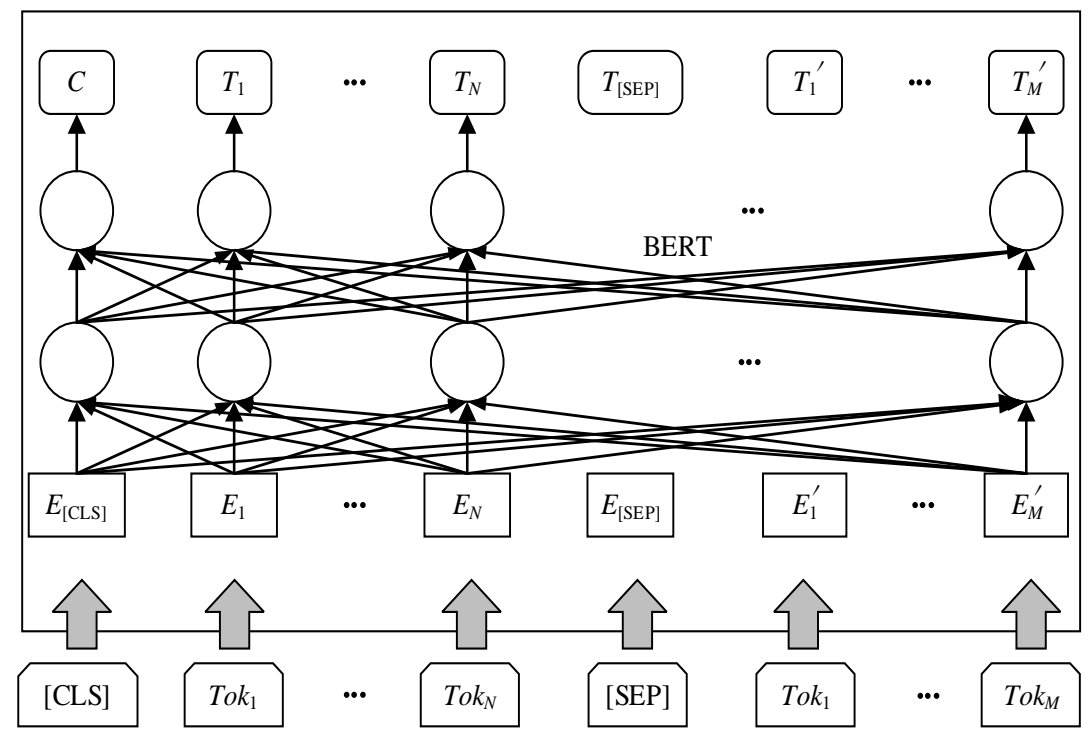

Fig.2 Framework of BERT model

The use of language models to assist natural language processing tasks has been widely discussed in the academic community [24]. There are two general approaches. 1 ) Feature-based, which refers to the use of intermediate results of the language model (language model word embeddings) as additional features to be introduced into the model of the original task, such as the ELMo model [25]; 2 ) based fine-tuning, refers to the use of a large number of corpus to train language models, and add a small number of neural network layers on the basis of the language model to accomplish specific tasks, using labeled corpus to supervise the training of new models, the process of the language model parameters are not fixed, such as GPT. the input of the above model as a left-to-right input of a sequence of text, or a combination of left-to-right input and right-to-left input training. However, BERT is a new model. However, BERT is a new pre-trained language model, i.e., a bi-directional encoded representation of the Transformer. Studies have shown that Bidirectionally trained language models have a deeper understanding of the context and are more efficient in extracting corpus features than unidirectional language models.

Since BERT can be used for a variety of natural language processing tasks and requires only a simple modification of the core model. After the features are fed into the model, they are passed to the word embedding layer, which includes performing token word embedding, sentence word embedding and positional word embedding. $Tok_i$ denotes the first $i$ Token, randomly masking part of the characters. $\boldsymbol{E}_i$ denotes the first $i$ the embedded vector of the Token. $\boldsymbol{T}_i$ denotes the first $i$ The feature vectors of each Token after BERT processing. BERT and the learnable weight matrix ($\boldsymbol{W}$) All parameters are fine-tuned to maximize the probability of correct classification. BERT generates a set of feature vectors based on the [CLS] flag $\boldsymbol{C}$, and combining it with $\boldsymbol{W}$ Multiplication. Initialize the model using the parameter weights of the pre-trained language model.

1.2.3 coding layer of fusion knowledge module

The coding layer of the fusion knowledge module is shown in Figure 2. Each candidate synonym $k \in Syn_{v_i}$ finds its corresponding vector $\boldsymbol{t}_k \in \mathbf{R}^{d\times 1}$ through the pre-trained word vector lookup table. In this paper, the attention mechanism with sentinel vector is used to dynamically decide whether to introduce external knowledge and distinguish which external knowledge is effective, so as to better measure the relationship between synonyms and contextual information. Sentinel vector

$$\boldsymbol{s_t} = \boldsymbol{\sigma}(\boldsymbol{W_b h_{t-1}^*} + \boldsymbol{U_b x_t})$$

Where: $\boldsymbol{W}_b$、$\boldsymbol{U}_b$ are the weight parameter matrices to be learned; $\boldsymbol{h}_{t-1}^*$ is the output vector of the last hidden state, and the text representation before the current input is saved as much as possible; $x_t$ is the currently input text vector.

Attention weight of synonym vector $\boldsymbol{t}_k$ and sentry vector $s_t$;

$$S(t_k h_i^*) = (t_{kb})^T tanh\ (W_t t_k + W_{ht} h_i^* + b_t),$$

$$S(s_t h_i^*) = (s_{kb})^T tanh\ (W_s s_t + W_{hs} h_i^* + b_s),$$

$$\alpha_{t_k} = \frac{exp\ \left(S(t_k h_i^*)\right)}{\sum_{k=1}^{num} S\left(t_k h_i^*\right) + S\left(s_t h_i^*\right)}.$$

$$\beta_t = \frac{exp\ \left(S(s_t h_i^*)\right)}{\sum_{k=1}^{num} S\left(t_k h_i^*\right) + S\left(s_t h_i^*\right)}.$$

Where: $\boldsymbol{W}_t$、$\boldsymbol{W}_{ht}$、$\boldsymbol{W}_s$、$\boldsymbol{W}_{hs}$ are the weight parameter matrices that the method needs to learn; $\boldsymbol{t}_{kb}$、$\boldsymbol{s}_{kb}$、$\boldsymbol{b}_t$、$\boldsymbol{b}_s$ are the weight parameter vectors that need to be learned by the method; $S(\cdot)$ is used to calculate the importance scores of synonym $\boldsymbol{t}_k$ and sentry vector $\boldsymbol{s}_t$ in the state of $\boldsymbol{h}_i^*$; $\alpha_{t_k}$、$\beta_t$ respectively represent the correlation between synonym $\boldsymbol{t}_k$ and sentry vector $\boldsymbol{s}_t$ and the current state $\boldsymbol{h}_i^*$.

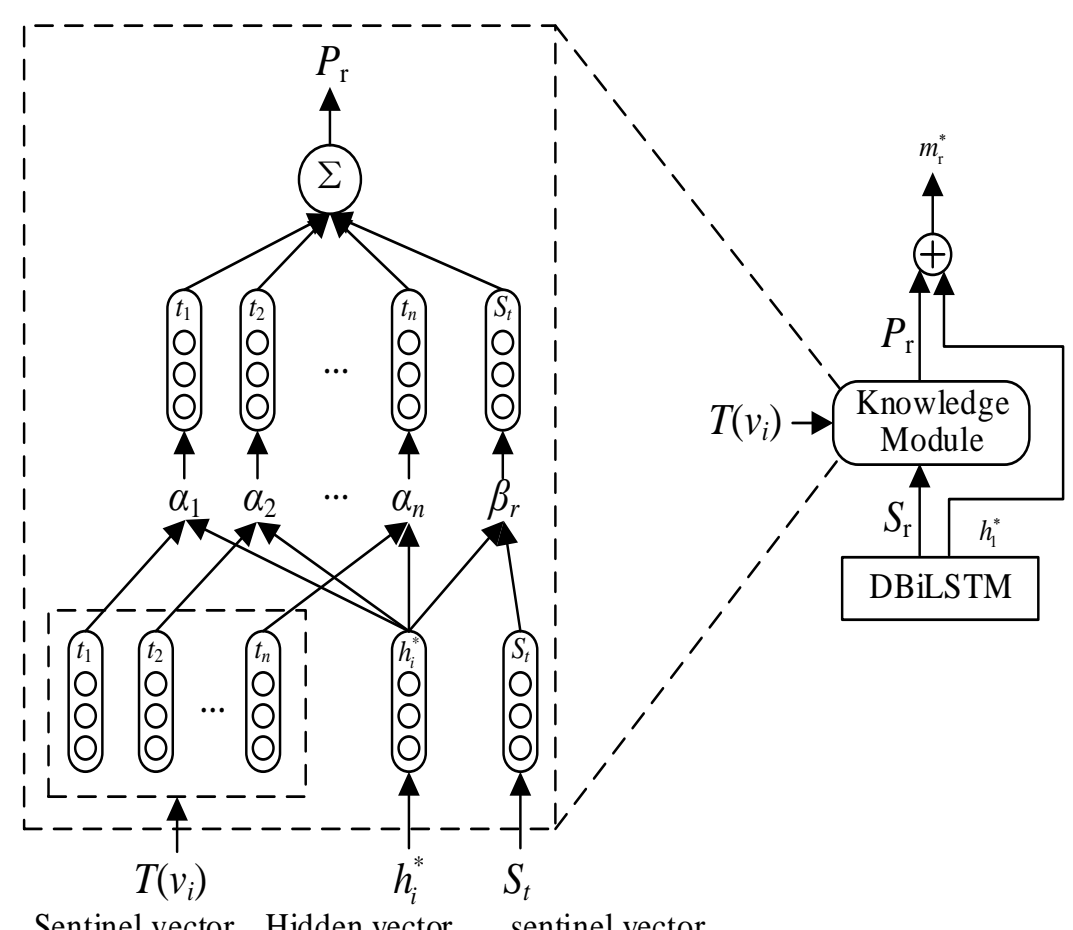

Fig.3 Embedded structure of knowledge graph

At the same time, in order to integrate the synonym information in the knowledge map into the coding layer, the knowledge state vector $\boldsymbol{p}_t$ is calculated by $\alpha_{t_k}$、$\beta_t$, and the knowledge state vector $\boldsymbol{p}_t$ is combined with the hidden layer vector $\boldsymbol{h}_i^*$ obtained by BERT, and the hidden layer vector $\boldsymbol{m}_i^*$ is called the knowledge perception state vector:

$$\boldsymbol{m}_i^* = \boldsymbol{p}_t + \boldsymbol{h}_i^*,$$
$$\boldsymbol{p}_t = \sum_{k \in \boldsymbol{Syn}_{v_i}} \boldsymbol{\alpha}_{t_k} \boldsymbol{t}_k + \boldsymbol{\beta}_t \boldsymbol{s}_t,$$
$$\sum_{k \in S_{yn_{v_i}}} \boldsymbol{\alpha}_{t_k} + \boldsymbol{\beta}_t = \boldsymbol{1}.$$

If $Syn_{v_i} = \emptyset$, let $\boldsymbol{p}_t = \boldsymbol{0}$, that is, the knowledge perception state vector is consistent with the hidden layer vector, where the knowledge state vector $\boldsymbol{p}_t$ and the hidden layer vector $\boldsymbol{h}_i^*$ have the same dimension.

Using knowledge-aware state vector $\boldsymbol{m}_i^*$ as memory block of text representation;

$$\boldsymbol{M}^* = \{\boldsymbol{m}_1^* \boldsymbol{m}_2^*, \cdots \boldsymbol{m}_t^*, \cdots \boldsymbol{m}_n^*\}$$

1.2.4 Textual representation incorporating positional attention information

A context word close to a perspective word is more important than a context word far away from the perspective word, and thus the positional weights at the $i$th word are as follows :

$$\boldsymbol{w}_i = \boldsymbol{1} - \frac{|\boldsymbol{i} - \boldsymbol{t}|}{\boldsymbol{t}_{max}}, \boldsymbol{u}_i = \frac{|\boldsymbol{i} - \boldsymbol{t}|}{\boldsymbol{t}_{max}}$$

where $t$ is the position of the perspective word, $t_{\max}$ is the total number of words in the input text, and $\mu_i$ is the relative positional distance between the context word and the perspective word.

If the perspective word consists of more than one word, calculating the distance $i - t$ between the context word and the perspective word determines whether the context word is to the left or to the right of the perspective word. If it is on the left, $f$ is the position of the leftmost word of the perspective word; if it is on the right, $f$ is the position of the rightmost word of the perspective word: thus, the distance between the context word and the perspective word is calculated. Combining the weights $w_i$ computed by the positional attention mechanism and the perceptual state vector $\boldsymbol{m}_i^*$, to generate a location-weighted memory block.

$$\boldsymbol{M} = \{\boldsymbol{m}_1, \boldsymbol{m}_2, \cdots \boldsymbol{m}_t, \cdots \boldsymbol{m}_n\},$$
$$\boldsymbol{m}_i = (\boldsymbol{w}_i \boldsymbol{m}_i^* \, \boldsymbol{\mu}_i) \in \boldsymbol{R}^{\vec{\boldsymbol{d}}_{l^+} + \boldsymbol{d}_{l+1}}.$$

### 1.2.5 Perspective-level affective features based on DCGRU

Lemma 1 The stable distribution of a diffusion process can be expressed as a weighted combination of infinite steps of random walks on an undirected graph of the following form.

$$\boldsymbol{P} = \sum_{k=0}^{\infty} \boldsymbol{\alpha}(1 - \boldsymbol{\alpha})^k (\boldsymbol{D}_O^{-1} \boldsymbol{W})^k$$

where, the $k$ represents the number of diffusion steps. In practice, it is common to use an explicit value to truncate the diffusion process and assign a trainable weight to each truncated step. In this paper, we refer to a diffusion process that is bi-directional, i.e., one that includes a reverse pass. This is added because two-way diffusion with backward passages provides greater flexibility for the model to capture the effects of upstream and downstream flows.

On the basis of Lemma 1, the graphical signal $X \in \mathrm{R}^{N \times P}$ and the filter $f_\theta$ Define a diffusion convolution operator, with a finite $K$ The weighted combination of two-step bidirectional random walks, as shown in Eq. (6), is

$$X_{:,p*g} f_\theta = \sum_{k=0}^{K-1} \left(\theta_{k,1} (\boldsymbol{D}_O^{-1} \boldsymbol{W})^k + \theta_{k,2} (\boldsymbol{D}_I^{-1} \boldsymbol{W}^{-1})^k\right),$$
$$p \in \{1, \cdots, P\}$$

Where, $\boldsymbol{\theta} \in \mathrm{R}^{K \times 2}$ is a parameter in the filter., $\boldsymbol{D}_O^{-1} \boldsymbol{W}$ and $\boldsymbol{D}_I^{-1} \boldsymbol{W}^{-1}$ represent the state transfer matrices of the forward and backward diffusion process, respectively. Although convolutional computation is computationally intensive, if the graph composed by the sensor network $G$ is a sparse graph, then it is possible to use a graph with complexity $O(K)$ The recursive sparse dense matrix multiplication of Eq. (6) completes the computation of Eq. (6) such that the total time complexity is $O(\boldsymbol{K}|\varepsilon|)$, which is much smaller than $O(N^2)$.

Using the convolution operation defined in Eq. (7) to construct a diffuse convolutional layer, Mapping $P$-dimensional features to $Q$ -dimensional output. The parameter tensor is defined as: $\boldsymbol{\Theta} \in \mathrm{R}^{Q \times P \times K \times 2} = [\boldsymbol{\theta}]_{q,p}$ , where $\boldsymbol{\Theta}_{q,p,\ldots:} \in \mathrm{R}^{K \times 2}$ denotes the number

The corresponding to the first $p$ The dimensional input and the first $q$ parameters of the convolutional filter for the dimensional output. Thus, the diffuse convolutional layer can be expressed as follows.

$$\boldsymbol{H}_{:,q} = \boldsymbol{\alpha} \left( \sum_{p=1}^{P} \boldsymbol{X}_{:,p*g} \boldsymbol{f}_{\Theta_{q,p,\cdots}} \right), q \in \{1, 2, \cdots, Q\}$$

where $\boldsymbol{X} \in \mathrm{R}^{N \times P}$ is the input data; the $\boldsymbol{H} \in \mathrm{R}^{N \times Q}$ is the output data after dimensionality reduction; $\alpha\alpha$ denotes the activation function, e.g. ReLU, Sigmoid function. The diffusion convolutional layer is capable of learning graph-structured data representations and can be trained using a stochastic gradient-based approach.

In order to further extract the intrinsic properties and characterize the temporal dependencies in the data, this paper uses GRU, a variant of recurrent neural network RNN, to simulate the relevance of each memory content, and uses the diffusion convolution operator mentioned above to replace the matrix multiplication in the GRU network, thus leading to a DCGRU network, as shown in Eq. (8) and Eq. (11):

$$\boldsymbol{r}_t = \boldsymbol{\sigma}\left(\Theta_r *_g [\boldsymbol{x}_t, \boldsymbol{h}_{t-1}] + \boldsymbol{b}_r\right)$$
$$\boldsymbol{z}_t = \boldsymbol{\sigma}\left(\Theta_u *_g [\boldsymbol{x}_t, \boldsymbol{h}_{t-1}] + \boldsymbol{b}_z\right)$$
$$\widetilde{\boldsymbol{h}}_t = \tanh\left(\boldsymbol{\Theta}_C *_g [\boldsymbol{x}_t, (\boldsymbol{r}_t) \odot \boldsymbol{h}_{t-1}] + \boldsymbol{b}_h\right)$$
$$\boldsymbol{h}_t = z_t \odot \boldsymbol{h}_{t-1} + (1 - z_t) \odot \widetilde{\boldsymbol{h}}_t$$

where, in the case of $\boldsymbol{x}_t$ and $\boldsymbol{h}_t$ denotes the input data and the output data of the hidden layer, the $r_t$ and $z_t$ means that at the moment $t$ The reset and update gates of $*_g$ corresponds to the diffusion convolution operator defined in Eq. (7), the $\Theta_r, \Theta_u$ and $\Theta_C$ denotes the parameters of the reset gate, update gate and activation state filter.The DCGRU network uses the BPTT algorithm to train the parameters. Compared with the traditional GRU model, DCGRU is differentiated by the following. The update gate $u^{(t)}$, reset gate $r^{(t)}$ and memory cells using a diffusion convolution operator, allowing the position dependence to be interpreted intuitively and computed efficiently.

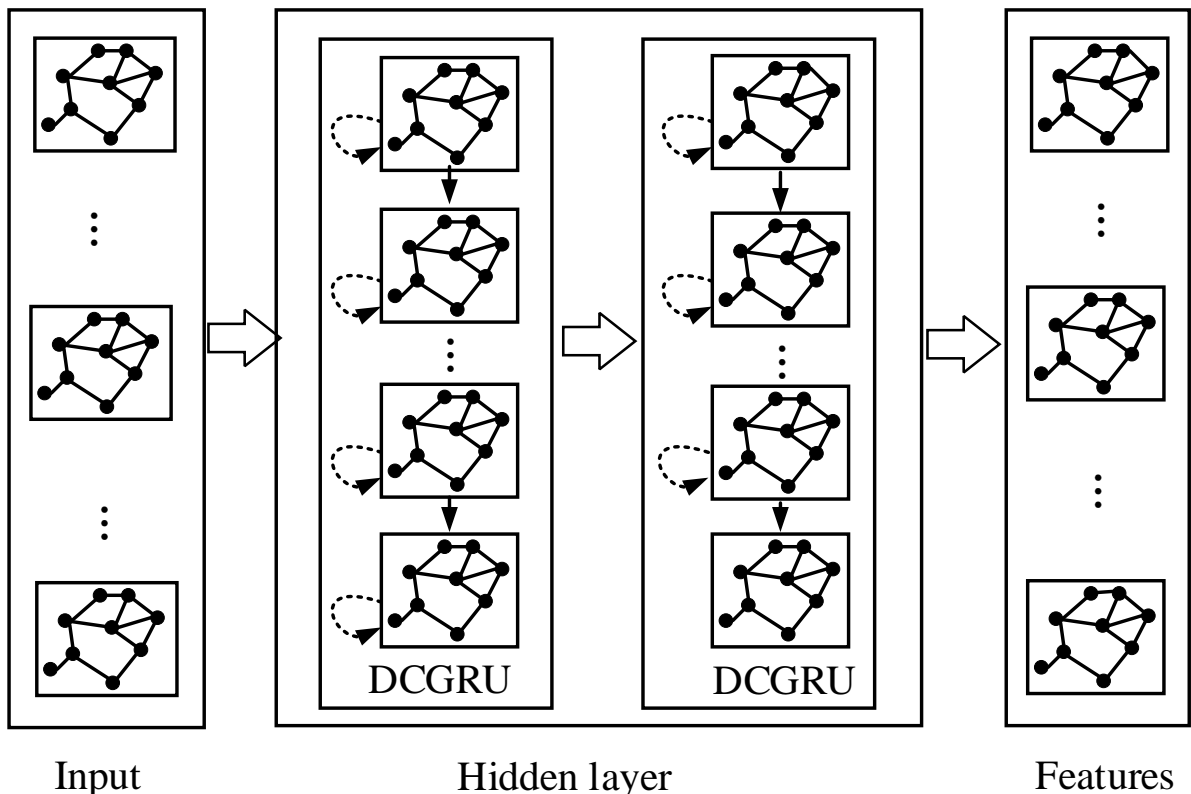

Fig.4 Structure of DCGRU

1.2.6 Output Layer

In this paper, the last output vector $\boldsymbol{e}_n$ is used as a feature, and it is inputted into the softmax layer for predicting the viewpoint-level text sentiment classification. The model is trained by minimizing the cross-entropy error with the following loss function.

$$\boldsymbol{loss} = \sum_{(\boldsymbol{sp}_a) \in \boldsymbol{T}} \sum_{\boldsymbol{c} \in \boldsymbol{y}} \boldsymbol{P}_{\boldsymbol{c}}^{\boldsymbol{g}}(\boldsymbol{sp}_a) \cdot \boldsymbol{ln}\left(\boldsymbol{P}_{\boldsymbol{c}}(\boldsymbol{sp}_a)\right) + \boldsymbol{\lambda} \parallel \boldsymbol{\theta} \parallel^2,$$

Among them, $T$ indicates that all sentences $y$ in the training set represent the tag set of emotion category, $(sp_a)$ indicates the training data pair composed of sentence and perspective, $P_c^g$ indicates the real emotion category tag of the text, $P_c$ indicates the predicted emotion category tag value, and the parameter set $\theta$ is adjusted during the model training. During the experiment, Adaptive estimates of lower-order moments (Adam) updating rules are used as the updating parameters of the optimization method, and $L_2$ regularization and random inactivation are used to reduce the influence of model over-fitting.

## 2 Experiment and result analysis

### 2.1 Experimental data set and experimental setup

In this paper, experiments are carried out on three public standard data sets: the data set of Restaurant and Laptop adopts the data set published by Semeval 2014 Task 4 [2]; The data set of Twitter domain adopts the data set collected by Dong et al [8]. These three data sets all contain three emotional polarity labels: Positive, Negative and Neutral. The details are shown in Table 1.

Table 1 Experimental datasets

| Name: Positive | | Positive | Negative | Neutral | Total |
| --- | --- | --- | --- | --- | --- |
| Laptop | Training set | 994 | 870 | 464 | 2328 |
| | Test set | 341 | 128 | 169 | 638 |
| Restaurant | Training set | 2164 | 807 | 637 | 3608 |
| | Test set | 728 | 196 | 196 | 1120 |
| | Training set | 1561 | 1560 | 3127 | 6248 |
| | Test set | 173 | 173 | 346 | 692 |

In this paper, WordNet is used to train the knowledge map embedding method by using the preprocessed data provided by Bordes et al [21], which includes 151,442 pairs of triples, 40,943 grammar sets and 18 relationships.

In the experiment, GloVe word vector lookup table is used to initialize word embedding, which is not updated during training. The word embedding dimension and hidden state dimension are both set to 300, the learning rate is set to 0.005, the weight coefficient $\lambda$ of $L_2$ regular term is 0.001, and the dropout coefficient is set to 0.5.

In this paper, Tensorflow is used to realize KG-ALSA and its methods are compared. The same input, word vector lookup table, word embedding dimension, hidden state dimension and optimizer are used.

### 2.2 Evaluation indicators

In this paper, the accuracy (ACC) [8] and the F1 value (Macro-F1, Mac-F1) under the macro average are used as experimental evaluation indicators to measure the overall experimental effect. The specific calculation formula is as follows:

$$Acc = \frac{total_true}{total_gold},$$

$$Mac\text{-}F1 = \frac{1}{class_num} \sum_{l=1}^{class_num} F1_l$$

$$F1_l = \frac{2 \cdot P_l \cdot R_l}{P_l + R_l},$$

$$P_l = \frac{n_true_l}{n_predict_l},\ R_l = \frac{n_true_l}{n_gold_l}$$

Where $n_true_l$ is the method to accurately predict the quantity of category $l$, $n_predict_l$ is the quantity of category $l$, $n_gold_l$ is the quantity actually belonging to category $l$, $P_l$ is the accuracy rate, and $R_l$ is the recall rate.

2.3 Comparison Methods

The comparison method selected is as follows.

1) SVM [4]. A series of features are extracted by SVM, and good results are achieved on Sem Eval2014 Task4.

2) LSTM [9]. Use LSTM to calculate the feature representation of the sentence. LSTM is a cyclic model, and the output vector of the hidden layer state corresponding to the last word is used as the sentence feature.

3) TD-LSTM [8]. Two LSTM networks are used, and then the representations are connected to predict the emotional polarity of perspective words.

4) Atae-LSTM [16]. Append perspective words after each word to enhance the influence of perspective words on text emotional polarity prediction, and use the LSTM network with attention mechanism to predict the emotional polarity of perspective words.

5) Ian [18]. Design attention mechanism, which can learn context representation and perspective word representation interactively.

6) Memnet [17], a multi-layer shared parameter calculation layer, each layer is a model based on contextual information and positional attention.

7) Ram [19]. The memory block constructed by MemNet is improved by DBiLSTM, and the threshold control unit is used to nonlinearly combine multiple attention output sentences.

8) Cabasc uses sentence-level attention mechanism to capture important information about a given perspective from the whole world, and considers words and their related order through contextual attention mechanism to solve the short-sighted problem of memory model.

9) Syntax-based mixed attention network [22]. Use the global attention mechanism to capture the rough information about the perspective, use the grammar-oriented local attention mechanism to observe the words close to the perspective grammatically.

2. 4 Effectiveness Comparison

In order to verify the effectiveness of the methods in this paper, the comparison of the ACC and MacF1 values of each method on three data sets are shown in Table 2. In the table, SVM, IAN, RAM, and SHAN are the experimental results obtained from related papers, and LSTM, TDLSTM, ATAELSTM, MemNet, and Cabasc are the results reproduced from related papers. It can be seen from Table 2 that the method in this paper has achieved the best results. The performance of KG-ALSA is better than that of Cabasc and SHAN, because different words may have different semantic information in different contexts. In KG-ALSA, the influence of synonym information in knowledge map on current information is measured by dynamic attention mechanism, which can better describe the semantic information of words in the current context, provide finer-grained features for emotion classification, and further improve the performance.

Table 2 Experimental results comparison of different algorithms on different datasets

| METHODOLOGY | Laptop | | Restaurant | | Twitter | |
|---|---|---|---|---|---|---|
| | Acc | *M* | Acc | Mac − F1 | Acc | MacF1 |
| SVM | 70.49 | - | 80.16 | - | - | - |
| LSTM | 66.77 | 61.78 | 74.29 | 62.58 | 66.33 | 63.37 |
| TD-LSTM | 67.71 | 60.25 | 75.36 | 64.48 | 69.51 | 67.13 |
| ATAE-LSTM | 68.50 | 61.52 | 77.32 | 64.99 | 68.93 | 67.22 |
| IAN | 72.10 | 63.10 | 78.60 | 67.40 | 68.80 | 66.60 |
| MemNet | 71.79 | 67.07 | 79.96 | 69.09 | 70.09 | 66.81 |
| RAM | 74.49 | 71.36 | 80.23 | 70.80 | 69.36 | 67.30 |
| Cabasc | 75.07 | 70.13 | 80.54 | 70.76 | 71.53 | 69.79 |
| SHAN | 74.64 | - | 81.02 | - | - | - |
| KG-ALSA | 79.64 | 75.43 | 86.25 | 77.74 | 76.13 | 75.27 |

Compared with LSTM and TD-LSTM SVM of neural network, the correct rate on Laptop and Restaurant datasets is 70.49% and 80.16% respectively. This is mainly because LSTM lacks consideration of the information of perspective words, and TD-LSTM cannot capture the interaction between perspective words and context. Moreover, the obtained features are mainly concentrated in the parts near the perspective words, which may ignore the long-distance emotional features and lead to poor performance, so feature engineering is helpful to improve performance, but it needs a lot of resources.

In the neural network based on attention mechanism, compared with MemNet, the accuracy of RAM on Laptop and Restaurant data sets is improved by 2.7% and 0.27% respectively, which is mainly because MemNet only uses one layer of attention mechanism to calculate, while RAM nonlinearly combines the results of different attention calculation layers for calculation. Therefore, the result is better than MemNet. Compared with RAM, the accuracy of CAASC on the three data sets is improved by 0.58%, 0.31% and 2.17% respectively, mainly because CAASC designs a sentence-level attention mechanism to capture the important information of words from a specific perspective from a global perspective, and measures the order and relationship between words and perspective words through a contextual attention mechanism. The memory content is customized for each perspective word, so the performance of Cabasc is better than that of RAM. SHAN makes full use of local information and all information based on syntax to realize the dynamic adjustment of attention weight, and achieves better performance than Cabasc on the Restaurant data set.

2.5 The influence of different embedding methods

The specific experimental results are shown in Table 3. As can be seen from the table, the performance improvement of the method is the most by using TransR. In Laptop, Restaurant and Twitter, the correct rates are 79.64%, 86.25% and 76.13% respectively. This is mainly because TransR pays attention to different attributes of entities for different relationships, and similar entities are close to each other in the entity space. However, the corresponding relational spaces are far away from each other in specific aspects. However, the translation embedding (TRANSE) [21] can only achieve good performance and scalability in a one-to-one simple relational model and a large-scale sparse knowledge base. Based on the hyperplane translation method (Translating on Hyperplanes, Transh) [23] Although it solves the shortcoming that TransE can't handle complex relations, entity vectors are projected into the semantic space of relations, which leads to their same dimensions and limits the expressive ability of the model. Therefore, TransR can map synonyms into low-dimensional space more accurately and obtain the best results.

Table 3 Effect of different knowledge graph embedding on performance of the proposed method

| Embedding method | Laptop | | Restaurant | | Twitter | |
|---|---|---|---|---|---|---|
| | Acc | MacF1 | Acc | MacF1 | Acc | MacF1 |
| TransH | 77.87 | 72.17 | 84.50 | 75.01 | 73.78 | 72.85 |
| TransE | 78.24 | 73.29 | 85.27 | 76.58 | 74.34 | 73.91 |
| TransR | 79.64 | 75.43 | 86.25 | 77.74 | 76.13 | 75.27 |

2.6 The influence of attention layers

This section analyzes the influence of different levels of attention calculation on the performance, and the specific results are shown in Figure 3. As can be seen from the figure, the method in this paper first increases and then decreases with the increase of the number of circulating attention levels, and the Acc and Mac-F1 values reach the best when the number of levels is 3, and the performance of using single-level attention calculation levels is always inferior to that of using multi-level attention calculation levels. This shows that in complex situations, Single-level attention calculation may not be able to effectively capture the key information of emotional features in the text. At the same time, the performance of the method does not increase with the increase of attention calculation layers. As can be seen from the figure, when the number of attention calculation layers is 3, the Acc and Mac-F1 values of the method in this paper get the optimal values on three data sets, but the performance of the method decreases after more than 3 layers. This is mainly because with the increase of the number of layers, the complexity of the method becomes more difficult to train and the generalization ability decreases.

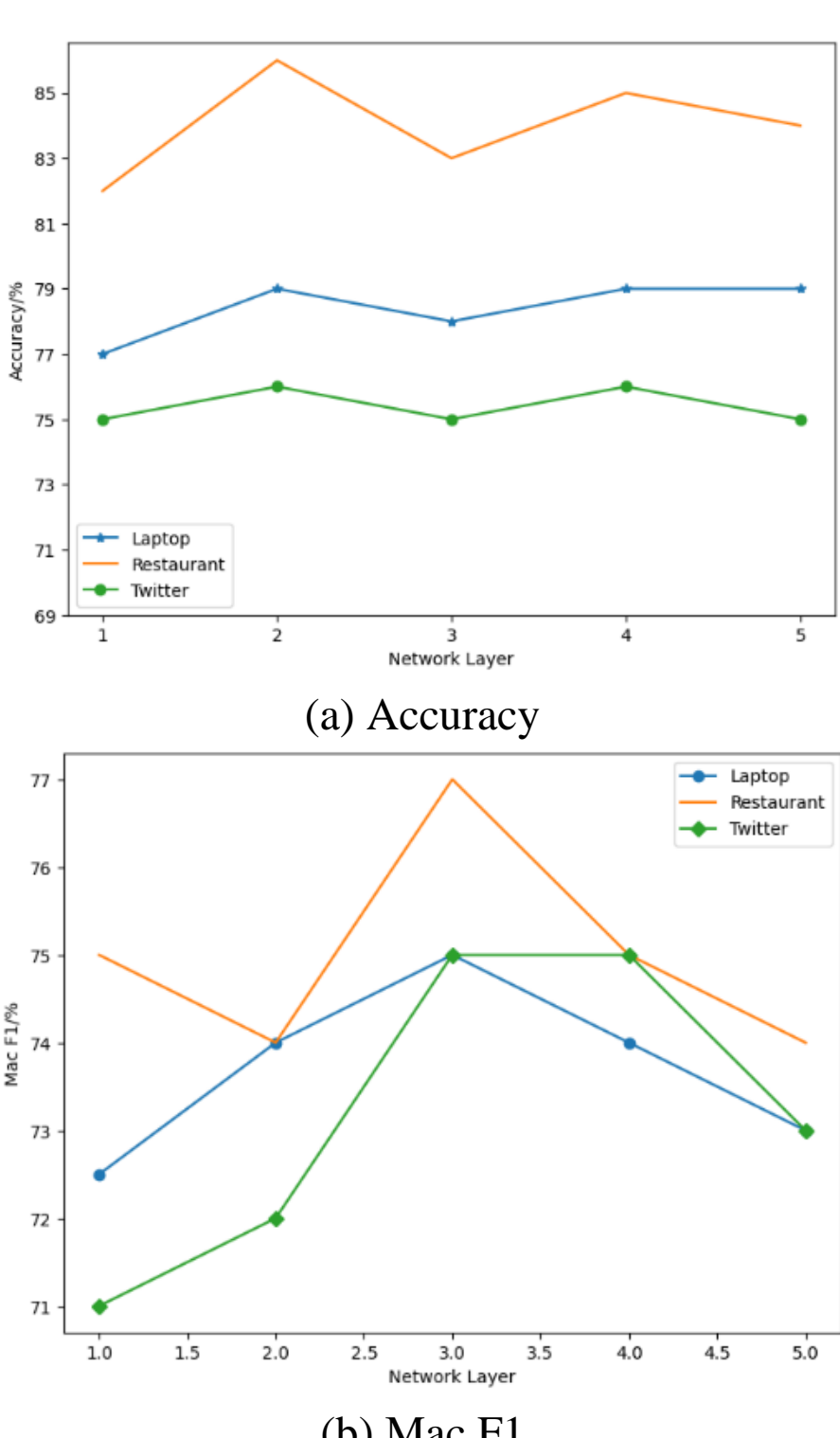

Fig.3 Effect of multiple attention layers

2.7 Effect of Attention Mechanism

In order to better illustrate the effect of the attention mechanism on text representation, the following example is chosen for attention visualization.

Example : Great food but the service was dreadful!
Translation : Great food but the service was dreadful!
Palindrome : food Polarity : Positive
Palindrome : service Polarity : Negative

The example contains two different perspective words, which have different emotional polarities under different perspective words. Figure 4(a) shows the visual effect of the attention weight of each word under the perspective word "food". It can be seen that the word "Great" has the highest attention weight after three levels of attention calculation. And correctly predict the emotional polarity of the text under the perspective word "food" as positive. (b) Visualize the attention weight of each word under the perspective word "service". At this time, the word "dreadful" has the highest attention weight, and correctly predict the emotional polarity of the text under the perspective word "service" as negative. This shows that the attention mechanism can effectively describe the importance of text words to emotional classification under different perspectives and better measure the contribution of words to emotional classification of the text.

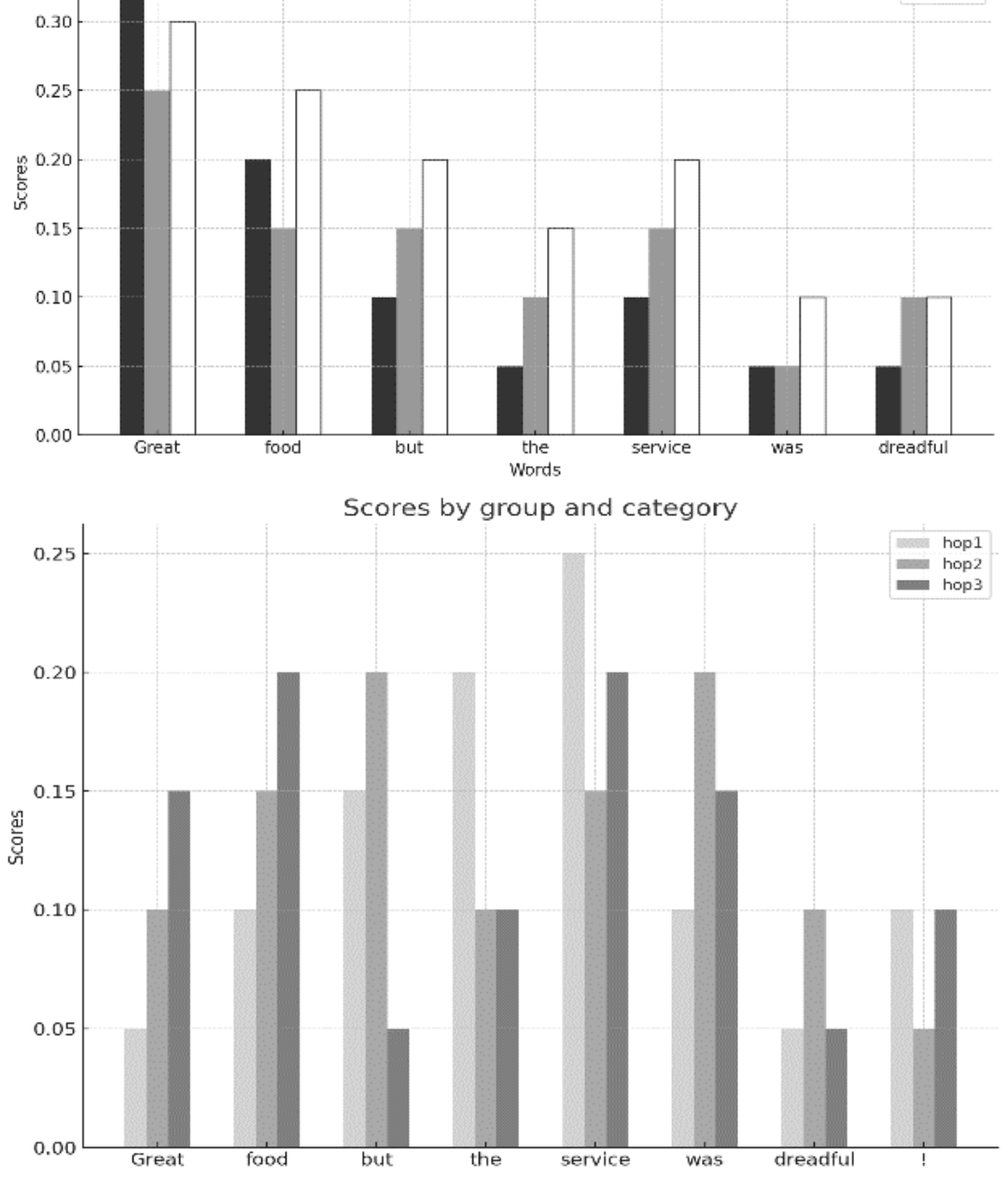

Fig.4 Attention weight of two aspect words

## 3 Conclusion

This paper presents a sentiment analysis approach that combines knowledge graphs and attention networks at the perspective level. It starts by encoding the input text's context using a BERT network. To handle word ambiguity across sentences, the model includes a knowledge graph through an attention mechanism. Then, it computes positional attention scores for each word relative to the perspective word. A threshold control unit combines previous round results with the current round's attention scores in a nonlinear manner. These resulting values become input features for the text and are used by the classifier for predictions. Experimental results show improved accuracy and macro-average F1 scores on three datasets. Future work may explore adding adversarial training to enhance the method's generalization and robustness.